\documentclass[letterpaper]{article} 
\usepackage{aaai25}  
\usepackage{times}  
\usepackage{helvet}  
\usepackage{courier}  
\usepackage[hyphens]{url}  
\usepackage{graphicx} 
\usepackage{natbib}  
\usepackage{caption} 
\usepackage{multirow}
\usepackage{algorithm}
\usepackage{algorithmic}
\newtheorem{definition}{Definition}
\newtheorem{theorem}{Theorem}
\newtheorem{proposition}[theorem]{Proposition}

\usepackage{dsfont}

\usepackage{amssymb}
\usepackage[table,xcdraw]{xcolor}
\usepackage{subcaption}
\usepackage{booktabs}

\usepackage{amsmath,amsfonts,bm}

\def\eqref#1{equation~\ref{#1}}

\def\1{\bm{1}}

\DeclareMathAlphabet{\mathsfit}{\encodingdefault}{\sfdefault}{m}{sl}
\SetMathAlphabet{\mathsfit}{bold}{\encodingdefault}{\sfdefault}{bx}{n}

\def\gB{{\mathcal{B}}}

\DeclareMathOperator*{\argmax}{arg\,max}

\usepackage{newfloat}
\usepackage{listings}
\DeclareCaptionStyle{ruled}{labelfont=normalfont,labelsep=colon,strut=off} 
\floatstyle{ruled}
\newfloat{listing}{tb}{lst}{}
\floatname{listing}{Listing}
\title{PROSAC: Provably Safe Certification for Machine Learning Models under Adversarial Attacks}
\author{
    Chen Feng\textsuperscript{\rm 1}, Ziquan Liu\textsuperscript{\rm 2}, Zhuo Zhi\textsuperscript{\rm 1}, Ilija Bogunovic\textsuperscript{\rm 1}, Carsten Gerner-Beuerle\textsuperscript{\rm 3}, Miguel Rodrigues\textsuperscript{\rm 4}
}
\affiliations{
    \textsuperscript{\rm 1}Department of Electronic and Electrical Engineering, University College London\\
    \textsuperscript{\rm 2}School of Electronic Engineering and Computer Science, Queen Mary University of London\\
    \textsuperscript{\rm 3}Faculty of Laws, University College London\\
    \textsuperscript{\rm 4}AI Centre, Department of Electronic and Electrical Engineering, University College London\\
    chen.feng@ucl.ac.uk, ziquan.liu@qmul.ac.uk, \{zhuo.zhi.21, i.bogunovic, c.gerner, m.rodrigues\}@ucl.ac.uk
}

\usepackage{bibentry}

\usepackage{cleveref}
\crefname{figure}{Fig.}{Figs.}       
\Crefname{figure}{Figure}{Figures}   

\crefname{table}{Tab.}{Tabs.}        
\Crefname{table}{Table}{Tables}      

\crefname{equation}{Eq.}{Eqs.}       
\Crefname{equation}{Equation}{Equations} 

\crefname{section}{Sec.}{Secs.}      
\Crefname{section}{Section}{Sections} 

\begin{document}

\maketitle

\begin{abstract}
It is widely known that state-of-the-art machine learning models, including vision and language models, can be seriously compromised by adversarial perturbations. It is therefore increasingly relevant to develop capabilities to certify their performance in the presence of the most effective adversarial attacks. Our paper offers a new approach to certify the performance of machine learning models in the presence of adversarial attacks with population level risk guarantees. In particular, we introduce the notion of $(\alpha,\zeta)$-safe machine learning model. We propose a hypothesis testing procedure, based on the availability of a calibration set, to derive statistical guarantees providing that the probability of declaring that the adversarial (population) risk of a machine learning model is less than $\alpha$ (i.e. the model is safe), while the model is in fact unsafe (i.e. the model adversarial population risk is higher than $\alpha$), is less than $\zeta$. We also propose Bayesian optimization algorithms to determine efficiently whether a machine learning model is $(\alpha,\zeta)$-safe in the presence of an adversarial attack, along with statistical guarantees. We apply our framework to a range of machine learning models - including various sizes of vision Transformer (ViT) and ResNet models - impaired by a variety of adversarial attacks, such as PGDAttack, MomentumAttack, GenAttack and BanditAttack, to illustrate the operation of our approach. Importantly, we show that ViT's are generally more robust to adversarial attacks than ResNets, and large models are generally more robust than smaller models. Our approach goes beyond existing empirical adversarial risk-based certification guarantees. It formulates rigorous (and provable) performance guarantees that can be used to satisfy regulatory requirements mandating the use of state-of-the-art technical tools.
\end{abstract}

\section{Introduction}
With the development of increasingly capable autonomous machine learning systems and their use in a range of domains from healthcare to banking and finance, education, and e-commerce, to name just a few, policy makers across the world are in the process of formulating detailed regulatory requirements that will apply to developers and operators of AI systems. The EU is at the forefront of the drive to regulate AI systems. Proposals for an EU AI Act, an AI Liability Directive
and an extension of the EU Product Liability Directive to AI systems and AI-enabled goods are at advanced stages of the legislative process. Other jurisdictions, too, pursue a variety of regulatory initiatives, and standard setters such as the National Institute of Standards and Technology in the United States and the Supreme Audit Institutions of Germany, the UK, and other countries have started work on more precise standards, including standards concerning the robustness of machine learning systems in the presence of adversarial attacks.

Regulatory frameworks adopted so far are mostly high-level, but those that establish more detailed requirements for AI systems to be put in service or for ongoing compliance, such as the EU AI Act, require an assessment of the performance of AI systems based on precise metrics. These metrics must include, among other things, an evaluation of the accuracy and resilience of a system in case of perturbations or unauthorised use.

It is thus important for those who deploy an AI system to have technical capabilities that allow a precise measurement of performance. However, developing certification procedures is not trivial due to the fact that state-of-the-art machine learning models are black-boxes that are poorly understood; furthermore, the standard train/validate/test paradigm often lacks rigorous quantifiable statistical guarantees and is therefore a poor certification instrument. Therefore, recent years have witnessed the introduction of various promising procedures, building on recent advances in statistics, that can be used to endow black-box  / complex state-of-the-art machine learning models with statistical guarantees \citep{bates2021distribution,angelopoulos2021learn,laufer-goldshtein2023}. For example, \citet{bates2021distribution} have proposed a framework to offer  rigorous distribution-free error control of machine learning models for a variety of tasks. \citet{angelopoulos2021learn} have proposed a procedure, the Learn-then-Test framework, that leverages multiple hypothesis testing techniques to calibrate machine learning models so that their predictions satisfy explicit, finite-sample statistical guarantees. Building on the Learn-then-Test framework, \citet{laufer-goldshtein2023} introduce a procedure to identify machine learning model risk-controlling configurations that also satisfy a variety of other objectives. Additionally, conformal prediction techniques have been proposed to quantify the reliability of the predictions of machine learning models, e.g. \citet{angelopoulos-bates2023}.

Our paper builds on this line of research to offer an approach -- Provably Safe Certification (PROSAC) -- to certify the robustness of a machine learning model under adversarial attacks \citep{bruna2014,Chakraborty2018} with population-level guarantees, thereby differing from existing approaches that are limited to the certification of the empirical risk such as \citet{cohen2019certified,wong2018provable}  (see Section 2). In particular, we build on hypothesis testing techniques akin to those in \citet{angelopoulos2021learn,laufer-goldshtein2023} to determine whether a model is robust against a specific adversarial attack. However, our approach differs from those in \citet{angelopoulos2021learn,laufer-goldshtein2023} because we aim to guarantee that a machine learning model is safe for any attacker hyper-parameter configuration, rather than for at least one such hyper-parameter configuration. PROSAC is then used to benchmark a wide variety of state-of-the-art machine learning models, such as vision Transformers (ViT) and ResNet models, against a number of adversarial attacks, such as PGDAttack~\cite{pgdattack}, MomentumAttack~\cite{momentumattack}, GenAttack~\cite{genattack} and BanditAttack~\cite{banditattack} in vision tasks.

\textbf{Contributions}: Our main contributions are as follows:
\begin{itemize}

\item We propose PROSAC, a new framework to certify whether a machine learning model is robust against a specific adversarial attack. Specifically, we propose a hypothesis testing procedure based on a notion of $(\alpha,\zeta)$ machine learning model safety, entailing (loosely) that the adversarial risk of a model is less than a (pre-specified) threshold $\alpha$ with a (pre-specified) probability higher than $\zeta$. 

\item We propose a Bayesian optimization algorithm –- concretely, the (Improved) GP-UCB algorithm –- to approximate the $p$-values associated with the underlying hypothesis testing problems, with a number of queries that scale much slower than the number of hyper-parameter configurations available to the attacker.

\item We also demonstrate that -- under a slightly more stringent testing procedure -- the proposed Bayesian optimization algorithm allows us to rigorously certify $(\alpha,\zeta)$-safety of a specific machine learning model in the presence of a specific adversarial attack.

\item Finally, we offer a series of experiments elaborating on $(\alpha,\zeta)$-safety of different machine learning models in the presence of different adversarial attacks. Notably, our framework reveals that ViTs appear to be more robust to adversarial perturbations than ResNets, and that large models appears to be more robust to adversarial perturbations than smaller models. 

\end{itemize}

\textbf{Organization}: Our paper is organized as follows: The following section briefly reviews related work. Section 3 presents the problem statement, including the notion of $(\alpha,\zeta)$ machine learning model safety under adversarial attacks. Section 4 presents our procedure to certify $(\alpha,\zeta)$ machine learning model safety. It describes the algorithm to certify $(\alpha,\zeta)$ machine learning model safety and presents associated guarantees. Section 5 offers experimental results to benchmark $(\alpha,\zeta)$-safety of various machine learning models under various attacks. Finally, we offer concluding remarks in Section 6. The proofs of the main technical results are relegated to the Supplementary Material.

\section{Related Works}
\subsubsection{Adversarial Robustness Certification}
Different approaches have been proposed to certify the adversarial robustness of machine learning models \citep{li2023sok}. For example, a) set propagation methods \citep{wong2018provable,wong2018scaling,gowal2018effectiveness,gowal2019scalable,zhang2019towards}; b) Lipschitz constant controlling methods \citep{hein2017formal,tsuzuku2018lipschitz,trockman2020orthogonalizing,leino2021globally,zhang2021boosting,xu2022lot}; and c) randomized smoothing techniques \citep{cohen2019certified,lecuyer2019certified,salman2019provably,carlini2023certified}. Set propagation approaches need access to the model architecture and parameters so that an input polytope can be propagated from the input layer to the output layer to produce an upper bound for the worse-case input perturbation. This approach however requires the model architecture to be able to propagate sets, e.g. \citep{wong2018provable} relies on ReLU activation functions.
~Lipschitz constant controlling approaches produce adversarial robustness certification by bounding local Lipschitz constants; however, these approaches are limited to certain model architectures such as LipConvnet \citep{singla2021skew}. In contrast, randomized smoothing (RS) represents a versatile certification methodology free from model architectural constraints or model parameter access.

\subsubsection{Other Certification Approaches} There are various other recent approaches to certify (audit) machine learning models in relation to issues including fairness or bias \citep{black2020fliptest,xue2020auditing,si2021testing,taskesen2021statistical,chugg2023auditing}. For example, \citet{black2020fliptest}, \citet{xue2020auditing}, \citet{taskesen2021statistical} and \citet{si2021testing} leverage hypothesis testing techniques -- coupled with optimal transport approaches -- to test whether a model discriminates against different demographic groups; \citet{chugg2023auditing} leverages recent advances in (sequential) hypothesis testing techniques -- the ``testing by betting'' framework -- to continuously test (monitor) whether a model is fair. Our certification framework also leverages hypothesis testing techniques, but the focus is on certifying the model adversarial robustness rather than model fairness.

\subsubsection{Distribution-free uncertainty quantification} Our certification framework builds on recent work on distribution-free risk quantification, e.g. \citet{bates2021distribution,angelopoulos2021learn}. In particular, \citet{bates2021distribution,angelopoulos2021learn} seek to identify model hyper-parameter configurations that offer a pre-specified level of risk control (under a variety of risk functions). See also similar follow-up work by \citet{laufer-goldshtein2023} and \citet{quach2023conformal}. Our proposed PROSAC framework departs from these existing approaches in that it seeks to offer risk guarantees for a machine learning model in the presence of an adversarial attack. Via the use of a GP-UCB algorithm, it seeks to ascertain the risk of a machine learning model in the presence of the worst-case attacker hyper-parameter configuration.

\section{Problem Statement}

\subsection{Adversarial Attack}
We consider how to certify the robustness of a (classification) machine learning model against specific adversarial attacks. We assume that we have access to a machine learning model $\mathcal{M}: \mathcal{X} \rightarrow \mathcal{Y}$ that maps features $X \in \mathcal{X}$ onto a (categorical) target $Y \in \mathcal{Y}$ where $(X,Y)$ are drawn from an unknown distribution $\mathcal{D}_{X,Y}$. We also assume that this machine learning model has already been optimized (trained) \textit{a priori} to solve a specific multi-class classification task using a given training set (hence, $\mathcal{Y} = \left\{1,2,\ldots,K\right\}$). We denote the corresponding positive loss function as $\mathcal{L}: \mathcal{Y} \times \mathcal{Y} \rightarrow \mathbb{R}_0^+$.

We consider that the machine learning model $\mathcal{M}$ is attacked by an adversarial attack $$\mathcal{A}_{\mathcal{M}, \mathcal{B}^q_{\epsilon}}: \mathcal{X} \times \mathcal{Y} \rightarrow \mathcal{X},$$ that given a pair $(X,Y) \in \mathcal{X} \times \mathcal{Y}$ -- (ideally) converts the original model input $X \in \mathcal{X}$ onto an adversarial one $\tilde X \in \mathcal{X}$ as follows:
\begin{equation}
\tilde X = \mathcal{A}_{\mathcal{M}, \mathcal{B}^q_{\epsilon}} (X,Y) = X + \argmax_{\delta\in\mathcal{B}^q_{\epsilon}}\mathcal{L}(\mathcal{M}(X+\delta), Y),
    \label{eqn:adv_attack}    
\end{equation}
with the intent of maximizing the loss for the given sample $(X,Y) \in \mathcal{X} \times \mathcal{Y}$, where $\gB_{\epsilon}^q$ is an $\ell_q$-norm bounded ball with radius $\epsilon$ (where $\epsilon$ measures the capability of the attacker, i.e., the attack budget).

However, practically, it is nontrivial to directly obtain the optimal adversarial sample -- calculating $\argmax_{\delta\in\mathcal{B}^q_{\epsilon}}\mathcal{L}(\mathcal{M}(X+\delta), Y)$ analytically with \cref{eqn:adv_attack} is often inaccessible. Most attackers often need to iterate and update based on the original sample, which requires manually setting the number of iterations and iteration step size, etc. Specifically, according to the accessibility of the model information, common attackers are divided into two categories: \textit{white-box} attacks, where the attacker has full access to the machine learning model, including its architecture/parameters/gradients, and \textit{black-box} attacks, where the attacker does not have full access to the machine learning model. 
In both conditions, we assume that the attacker draws its hyper-parameter configuration (for example, iteration steps) $\lambda$ from a (finite) set of hyper-parameter configurations $\Lambda$, where each hyper-parameter configuration is $d$-dimensional i.e. $\lambda \in \mathbb{R}^d$. 

Moreover, in general, the various attacks are still stochastic. Given fixed attack hyper-parameters $\lambda$, the white-box and black-box attacks do not deliver a deterministic perturbation $\tilde{\delta}$ given fixed sample $(X,Y)$ but rather random perturbations, because the attacks depend on other random variables. For example, the white-box PGDAttack~\citep{pgdattack} depends on the random initialization.
Denoted as $Z$ such remaining randomness in the attack, therefore, the adversarial attacks\footnote{Note that, we do not consider the attack budget, $\epsilon$, to be a hyper-parameter since it would not be possible to control the risk where the adversary has the ability to choose any attack budget $\epsilon \in (0,\infty)$. We also do not consider the attack norm to be a hyper-parameter.} can be represented as 
\[\mathcal{A}_{\mathcal{M},\mathcal{B}_\epsilon^q,\lambda}:(\mathcal{X}\times\mathcal{Y})\times\mathcal{Z}\rightarrow \mathcal{X},\] and the corresponding adversarial sample as:
\begin{equation}
\tilde{X} = \mathcal{A}_{\mathcal{M},\mathcal{B}_\epsilon^q,\lambda}(X, Y, Z),
\end{equation}


\subsection{Model Safety}
Given an adversarial attack $\mathcal{A}_{\mathcal{M},\mathcal{B}_\epsilon^q,\lambda}$, we can consequently characterize the safety of a machine learning model using two quantities: the \textit{adversarial risk} and the \textit{max adversarial risk}. We define the adversarial (population) risk induced by an attack $\mathcal{A}_{\mathcal{M},\mathcal{B}_\epsilon^q,\lambda}$ on a model $\mathcal{M}$ as follows:
\begin{equation}  \label{eqn:adversarial risk}
\begin{split}
&\mathcal{R}_{\mathcal{A}_{\mathcal{M},\mathcal{B}_\epsilon^q,\lambda}} (\mathcal{M}) :=\mathbb{E}_{(X,Y,Z)\sim \mathcal{D}_{X,Y} \times \mathcal{D}_{Z}} \big\{ R_{\mathcal{M}} \big\},   
\end{split}
\end{equation}
where $$R_{\mathcal{M}}=\mathds{1}[ \mathcal{M} (\mathcal{A}_{\mathcal{M},\mathcal{B}_\epsilon^q,\lambda} (X,Y,Z)) \neq Y] \cdot \mathds{1}[ \mathcal{M} (X) = Y],$$
and we define the max adversarial (population) risk induced by an attack $\mathcal{A}_{\mathcal{M},\mathcal{B}_\epsilon^q,\lambda}$ on a model $\mathcal{M}$ independently of how the attacker chooses its hyper-parameters as follows:
\begin{equation} \label{eqn:max adversarial risk}
\begin{split}\mathcal{R}^{\star}_{\mathcal{A}_{\mathcal{M},\mathcal{B}_\epsilon^q}} (\mathcal{M})=\max_{\lambda}\mathcal{R}_{\mathcal{A}_{\mathcal{M},\mathcal{B}_\epsilon^q,\lambda}} (\mathcal{M}),
\end{split}
\end{equation}
 where we use the 0-1 loss to measure the per-sample loss\footnote{This work concentrates primarily on classification problems with 0-1 loss. However, our work readily extends to other losses subject to some modifications.}.  Note that the adversarial (population) risk characterizes the performance of the machine learning model for a specific attack with a given budget / norm and a fixed hyper-parameter configuration, whereas the max adversarial (population) risk characterizes the performance of the machine learning model for an attack with a given budget / norm, independently of how the attacker chooses its hyper-parameter configuration.


Our main goal is to determine whether a machine learning model is safe by establishing whether the max (adversarial) population risk is below some threshold with high probability.

\begin{definition} \label{def:model safety}
    ($(\alpha,\zeta)$-Model Safety) Fix $0 \leq \alpha \leq 1$, $0 \leq \zeta \leq 1$. Then, we say that a machine learning model $\mathcal{M}$ is $(\alpha,\zeta)$-safe under an adversarial attack $\mathcal{A}_{\mathcal{M},\mathcal{B}_\epsilon^q,\lambda}$ with fixed budget $\epsilon$ and $\ell_q$-norm, and for all attack hyper-parameters $\lambda\in\Lambda$, provided that
        \begin{equation}
        \begin{split}
\mathbb{P} \left({\sf reject}~ \mathcal{R}^*> \alpha|~  \mathcal{R}^*> \alpha {\sf~is~true} \right) \leq \zeta.
        \end{split}
    \end{equation}
Here, $\mathcal{R}^* \triangleq \mathcal{R}^*_{\mathcal{A}_{\mathcal{M},\mathcal{B}_\epsilon^q,\lambda}} (\mathcal{M}) $.
\end{definition}

We will see in the following that this entails formulating a hypothesis testing problem where the null hypothesis is associated with a max adversarial risk higher than $\alpha$. Therefore, $(\alpha,\zeta)$-model safety means that the probability of declaring that the max adversarial risk of a model is less than $\alpha$ when it is in fact higher than $\alpha$ is smaller than $\zeta$, or, more loosely speaking,
a model max adversarial risk is less than $\alpha$ with a probability higher than $1-\zeta$.

\section{Certification Procedure}

We now describe our proposed certification approach allowing us to establish $(\alpha,\zeta)$- safety of a machine learning model in the presence of an adversarial attack.
~We will omit the dependency of adversarial risk on the model, the attack, and the attack parameters in order to simplify notation. We will also omit the fact that the attack depends on the model, its budget / norm, and the hyper-parameters.

\subsection{Procedure}

Our procedure is related to, but also departs from, a recent line of research concerning risk control in machine learning models, pursued by \citet{bates2021distribution,angelopoulos2021learn} and \citet{laufer-goldshtein2023} (see also references therein). In particular, \citet{bates2021distribution,angelopoulos2021learn} and \citet{laufer-goldshtein2023} offer a methodology to identify a set of model hyper-parameter configurations that control the (statistical) risk of a machine learning model. However, we are not interested in determining a set of attacker hyper-parameters guaranteeing risk control, but rather in guaranteeing risk control independently of how an attacker chooses the hyper-parameters (since a user cannot control the choice of hyper-parameters).



We fix the machine learning model $\mathcal{M}$, the adversarial attack $\mathcal{A}$, the adversarial attack budget $\epsilon$, and the adversarial attack $\ell_q$-norm. We leverage -- in line with \citet{bates2021distribution,angelopoulos2021learn,laufer-goldshtein2023} -- access to a calibration set $\mathcal{S} = \left\{ (X_1,Y_1), (X_2,Y_2),\ldots,(X_n,Y_n) \right\}$ (independent of any training set) where the samples $(X_i,Y_i)$ are drawn i.i.d. from the distribution $\mathcal{D}_{X,Y}$ to construct our certification procedure. 


Our certification procedure then involves the following sequence of steps:

\begin{itemize}
    

    \item First, we set up a hypothesis testing problem where the null hypothesis is $\mathcal{H}_0 : \mathcal{R}^* > \alpha$ or, equivalently, $\mathcal{H}_0 : \exists~\lambda\in\Lambda, \mathcal{R}_\lambda > \alpha$, where $\mathcal{R}^*$ represents the max adversarial risk in \cref{eqn:max adversarial risk} and $\mathcal{R}_{\lambda}$ represents the adversarial risk in \cref{eqn:adversarial risk} that depends on the attacker hyper-parameters $\lambda \in \Lambda$. 


    \item Second, we leverage the calibration set (plus another set with a number of instances / objects characterizing the randomness of the attack) to determine a finite-sample $p$-value $p^*$ that can be used to reject the null hypothesis $\mathcal{H}_0 : \mathcal{R}^* > \alpha$ or, equivalently, $\mathcal{H}_0 : \exists~\lambda\in\Lambda, \mathcal{R}_{\lambda} > \alpha$.


    \item Finally, we reject the null hypothesis $\mathcal{H}_0 : \mathcal{R}^* > \alpha$ or, equivalently, $\mathcal{H}_0 : \exists~\lambda\in\Lambda, \mathcal{R}_{\lambda} > \alpha$ provided that the $p$-value $p^*$ is less than $\zeta$.
    
\end{itemize}



This procedure allows us to establish $(\alpha,\zeta)$- safety of the machine learning model $\mathcal{M}$ in the presence of an adversarial attack $\mathcal{A}$, in accordance with Definition \ref{def:model safety}.


\begin{proposition}\label{prop: p-value}
Let $p^*$ be a p-value associated with the hypothesis testing problem where the null hypothesis is $\mathcal{H}_0 : \mathcal{R}^* > \alpha$ or, equivalently, $\mathcal{H}_0 : \exists~\lambda\in\Lambda, \mathcal{R}_{\lambda} > \alpha$. It follows immediately that the machine learning model is $(\alpha,\zeta)$- safe, i.e.
    \begin{align}
\mathbb{P} \left({\sf reject}~ \mathcal{R}^* > \alpha ~\big|~ \mathcal{R}^* > \alpha {\sf~is~true} \right) \leq \zeta,
    \end{align}
    provided that the null hypothesis is rejected if and only if $p^* \leq \zeta$.
\end{proposition}

We next show how to derive a $p$-value for our hypothesis testing problem where $\mathcal{H}_0 : \exists~\lambda\in\Lambda, \mathcal{R}_{\lambda} > \alpha$ from the $p$-values for the hypothesis testing problems where $\mathcal{H}_0 : \mathcal{R}_{\lambda} > \alpha$, for all $\lambda\in\Lambda$ (see also \citealp{laufer-goldshtein2023}). \footnote{Note the difference between the hypothesis testing problems. The problem with the null $\mathcal{H}_0 : \exists~\lambda\in\Lambda, \mathcal{R}_{\lambda} > \alpha$ tests whether the max adversarial risk is above $\alpha$ independently of the choice of hyper-parameters associated with the attack, whereas the hypothesis testing problem with the null $\mathcal{H}_0 : \mathcal{R}_{\lambda} > \alpha$ tests whether the risk is above $\alpha$ for a particular choice of hyper-parameters associated with the attack.}


\begin{theorem} \label{thm:p-value1}
    If $p (\lambda)$ is a $p$-value associated with the null $\mathcal{H}_0 : \mathcal{R}_{\lambda} > \alpha$ then $p^* = \max_{\lambda \in \Lambda} p (\lambda)$ is a p-value associated with the null hypothesis $\mathcal{H}_0 : \exists~\lambda\in\Lambda,  \mathcal{R}_{\lambda} > \alpha$.
\end{theorem}


Therefore, building on Theorem \ref{thm:p-value1}, we can immediately determine a $p$-value for our hypothesis testing problem. 

\begin{theorem} \label{thm:p-value2}
  A (super-uniform) $p$-value associated with the null hypothesis $\mathcal{H}_0 : \exists~\lambda \in \Lambda,  \mathcal{R}_{\lambda} > \alpha$ is given by~\cite{bates2021distribution, angelopoulos2021learn}:
    \begin{equation}
    \begin{split}
        p^* = \max_{\lambda \in \Lambda}\ \min \Big\{ &\exp \left( -n \cdot h_1 ( \hat{\mathcal{R}} (\lambda) ; \alpha ) \right) , \\ & e \cdot \mathbb{P} \left( {\sf Bin} (n,\alpha) \leq \left \lceil n \cdot \hat{\mathcal{R}} (\lambda) \right \rceil \right)  \Big\}, \label{eqn:p-value}
        \end{split}
    \end{equation}
    where $\hat{\mathcal{R}} (\lambda)$ represents the adversarial empirical risk induced by the attack $\mathcal{A}_{\lambda}$ on model $\mathcal{M}$ given a specific hyper-parameter configuration $\lambda \in \Lambda$ i.e.
    \begin{equation}\label{eq:attacksuccessrate}
    \begin{split}
    \hat{\mathcal{R}} (\lambda,\mathcal{S},\mathcal{Z}) &= \frac{1}{n} \sum_{i=1}^{n} \mathds{1}[ \mathcal{M} (\mathcal{A}_{\lambda} (X_i,Y_i,Z_i)) \neq Y_i] \\&\qquad\qquad\cdot \mathds{1}[ \mathcal{M} (X_i) = Y_i],
        \end{split}
    \end{equation}
    where $\mathcal{S} = \big\{(X_1,Y_1),\dots,(X_n,Y_n)\big\}$ is the set containing the calibration data, $\mathcal{Z} = \left\{Z_1,\dots,Z_n\right\}$ is a set containing a series of random objects that capture the randomness of the attack, and $h_1 (a,b) = a \cdot \log (a/b) + (1-a) \cdot \log ((1-a)/(1-b))$. 
\end{theorem}

\subsection{Algorithm and Associated Guarantees}

Our procedure to establish $(\alpha,\zeta)$-safety of a machine learning model $\mathcal{M}$ in the presence of an adversarial attack $\mathcal{A}$, in accordance with Definition~\ref{def:model safety}, relies on the ability to approximate the $p$-value associated with the null hypothesis $\mathcal{H}_0 : \exists~\lambda\in\Lambda, \mathcal{R}_{\lambda} > \alpha$ as per Theorem \ref{thm:p-value2}. However, this involves solving a complex optimization problem concerning the maximization of a function (a Hoeffding-Bentkus $p$-value \citep{bates2021distribution,angelopoulos2021learn}) over the set of attacker hyper-parameter configurations. We therefore propose to adopt a Bayesian optimization (BO) procedure, based on the established Gaussian Process Upper Confidence Bound (GP-UCB) algorithm \citep{srinivas2010gaussian}, which can be used to search effectively over the set of hyper-parameter configurations of the attack in order to identify the configuration leading to the highest $p$-value\footnote{We require a sample-efficient optimization method since the evaluation of the $p$-value involves computation of the empirical risk of the model subject to the attack for each individual attack hyper-parameter configuration; this is very time-consuming for complex models used in our experiments.}. 

Algorithm~\ref{algo:ucb} summarizes the GP-UCB procedure used to search for the attack hyperparameters that solve \cref{eqn:p-value}. 
~The algorithm first ingests the attacker hyper-parameter grid configuration, the Gaussian process mean function, and the Gaussian process prior covariance (kernel) function. We select the kernel to be Matern kernel \citep{genton2001classes}. At round $t$, the algorithm determines the hyper-parameter configuration $\lambda_t \in \Lambda$ that maximizes the upper confidence bound. The algorithm then determines a $p$-value $\hat{p}_t$ corresponding to the sum of $p (\lambda_t)$ plus some i.i.d. zero-mean Gaussian noise $\nu_t$ (where $p (\lambda_t)$ is derived from \cref{eqn:p-value}, and the algorithm performs Bayesian updating to obtain a new GP mean function $\mu_t$ and covariance function $\sigma_t$. The algorithm finally delivers the $p$-value estimate after $T$ rounds. We choose $\beta_t$ to be 0.1 with hyper-parameter search from $\beta$=\{0.01,0.1,1.0\}.

\begin{algorithm}[t]
\caption{GP-UCB for hyperparameter optimization}\label{algo:ucb}
\begin{algorithmic}
\STATE \textbf{Input:} Hyper-parameter configuration grid $\Lambda$. Gaussian Process prior mean $\mu_0 = 0$; Gaussian Process prior covariance $\sigma_0 = k$ where $k$ corresponds to the kernel function. 
\FOR{t = 1, 2, 3 \ldots T}
\STATE Compute $\lambda_t = \argmax\limits_{\lambda\in \Lambda} \mu_{t-1}(\lambda) + \beta_t \sigma_{t-1}(\lambda)$.
\STATE Compute $\hat p_t = p(\lambda_t)+\nu_t$
\STATE Perform Bayesian update to obtain new GP mean $\mu_t$ and covariance $\sigma_t$ using the sampled points $(\lambda_t,\hat p_t)$.
\ENDFOR
\RETURN $\hat{p}_T = 1/T \sum_{t=1}^T \hat{p}_t$ 
\end{algorithmic}
\end{algorithm}

The following theorem shows that we can establish $(\alpha,\zeta)$-safety of the machine learning model $\mathcal{M}$ in the presence of an adversarial attack $\mathcal{A}$ (in accordance with Definition \ref{def:model safety}) by relying on Algorithm~\ref{algo:ucb}. In particular, in view of the fact that the GP-UCB procedure in Algorithm~\ref{algo:ucb} delivers a $p$-value estimate that is close to the true $p$-value with probability $(1-\delta)$, where $0 < \delta < 1$ (see guarantees in \citep{srinivas2010gaussian}), the hypothesis testing procedure underlying Theorem \ref{thm:model safety with gp ucb} compares the GP-UCB $p$-value estimate $\hat{p}_T$ to a more conservative threshold $\zeta' < \zeta$, rather than $\zeta$, where
\begin{equation} \label{threshold}
\zeta' = \zeta - \mathcal{O} \left( B \sqrt{ \gamma_T/T} + \sqrt{\gamma_T \left( \gamma_T + \log (1/\delta) \right)/T} \right) - \delta,    
\end{equation}
where the value $B$ bounds the smoothness of the $p$-value function
,  $\gamma_T$ corresponds to the maximum information gain at round $T$
, and $T$ is the number of GP-UCB rounds. According to \cref{threshold}, with probability nearly 1, we can expect $\zeta' \rightarrow \zeta$ with $T \rightarrow \infty$. The Supplementary Material
demonstrates that this more conservative testing procedure is sufficient to retain the $(\alpha,\zeta)$ safety guarantees in Definition \ref{def:model safety}.

\begin{figure*}[htbp]
    \centering
    \begin{subfigure}[h]{0.33\textwidth}
        \centering
        \includegraphics[width=\textwidth]{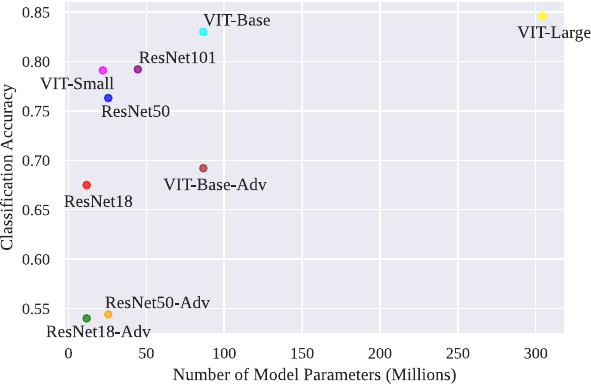}
        \caption{ImageNet Top-1 classification accuracy of different models.}
    \end{subfigure}
    \hfill
    \begin{subfigure}[h]{0.33\textwidth}
        \centering
        \includegraphics[width=\textwidth]{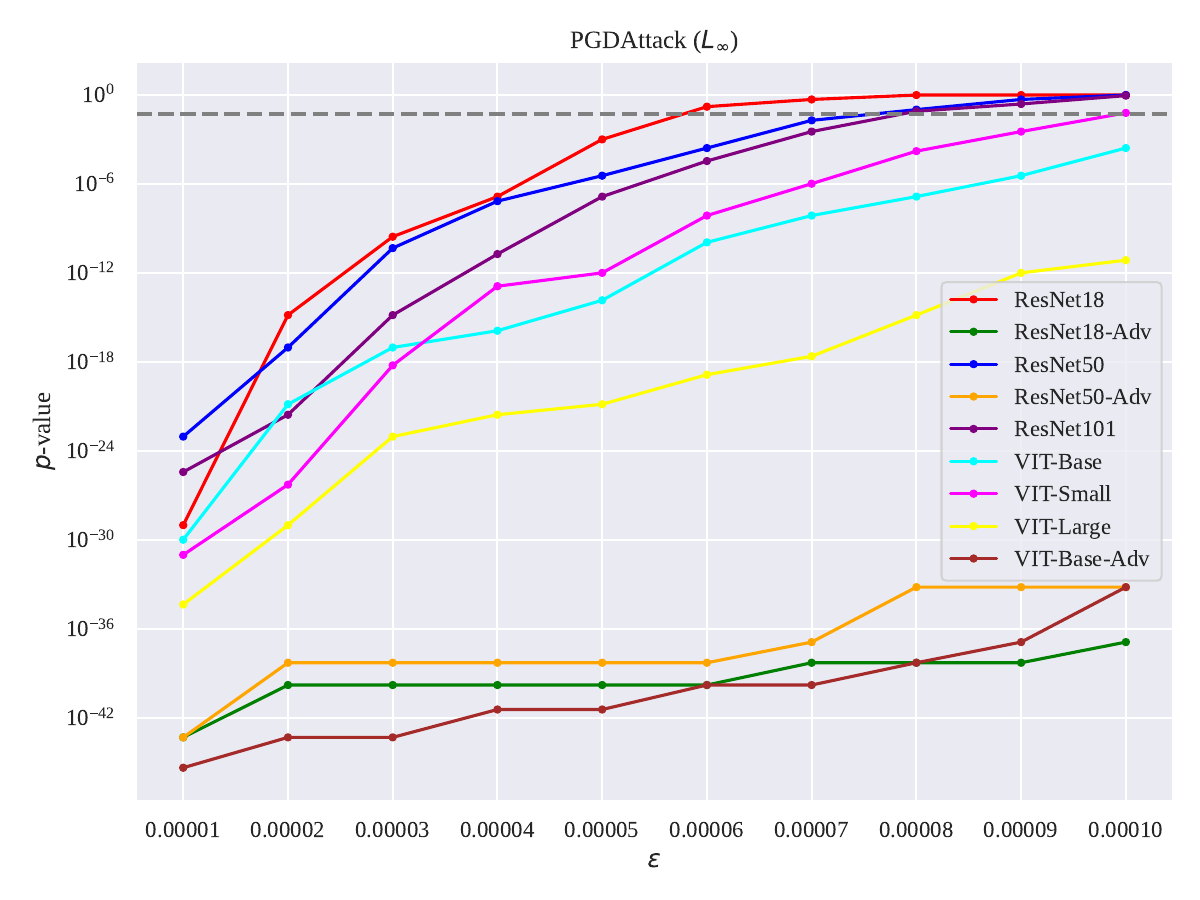}
        \caption{PGDAttack with $L_{\infty}$ norm.}
    \end{subfigure}
    \hfill
    \begin{subfigure}[h]{0.33\textwidth}
        \centering
        \includegraphics[width=\textwidth]{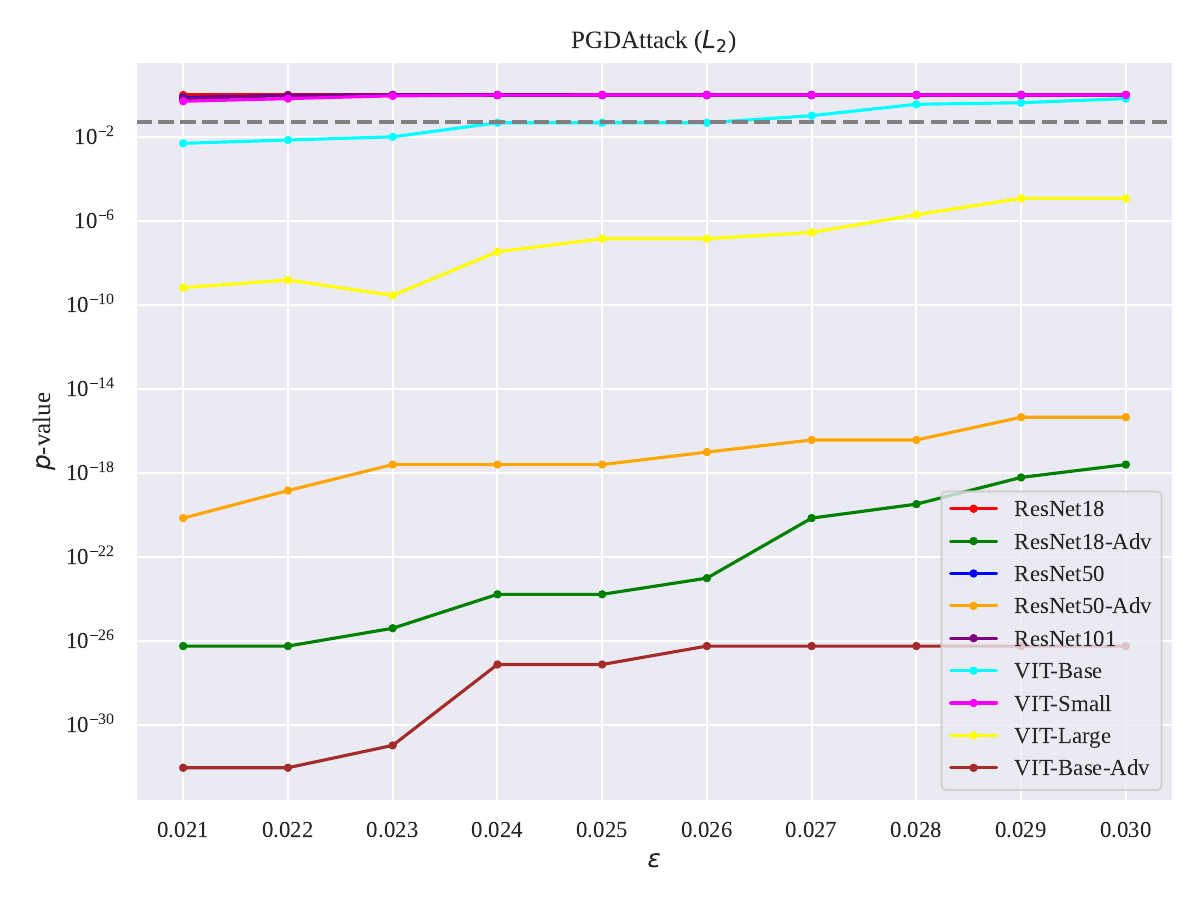}
        \caption{PGDAttack with $L_2$ norm.}
    \end{subfigure}
    
    \vskip\baselineskip
    
    \begin{subfigure}[h]{0.33\textwidth}
        \centering
        \includegraphics[width=\textwidth]{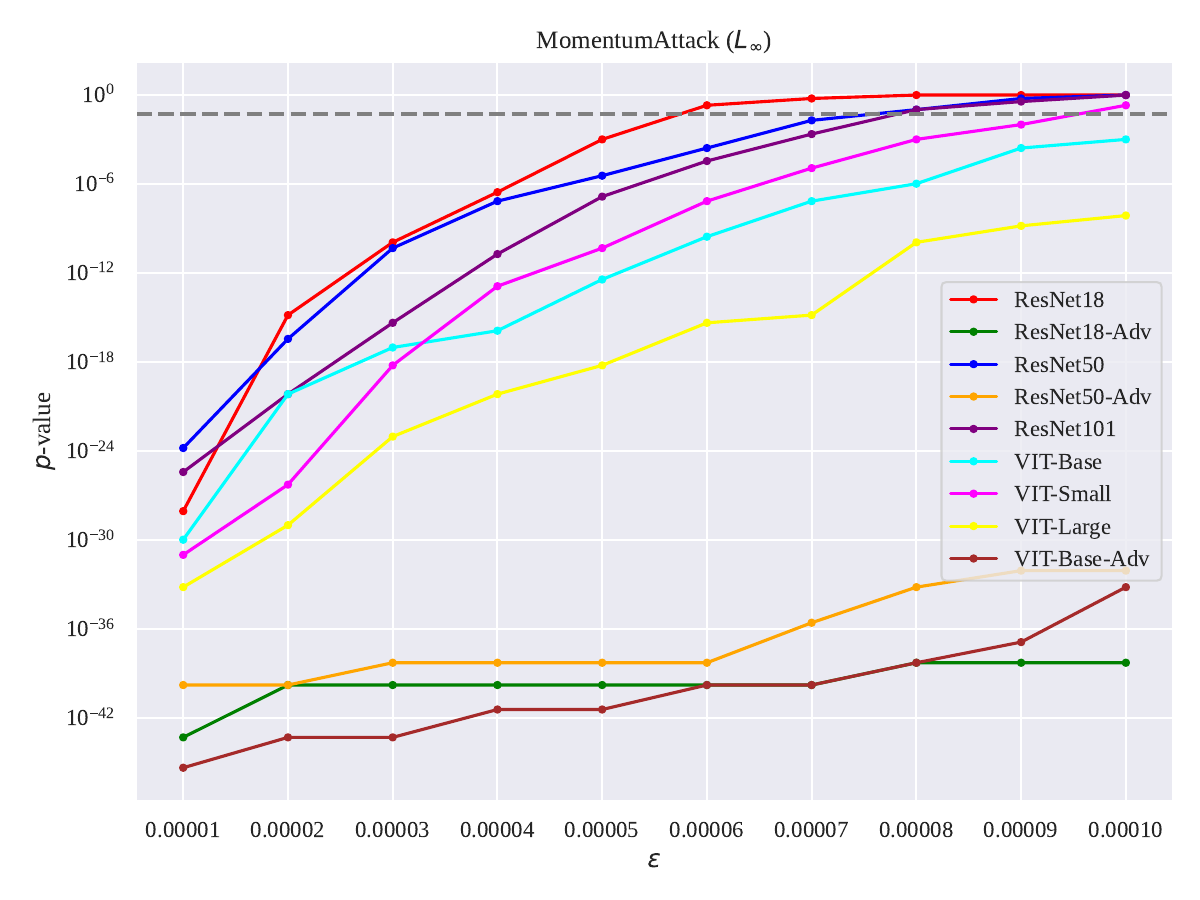}
        \caption{MomentumAttack with $L_{\infty}$ norm.}
    \end{subfigure}
    \hfill
    \begin{subfigure}[h]{0.33\textwidth}
        \centering
        \includegraphics[width=\textwidth]{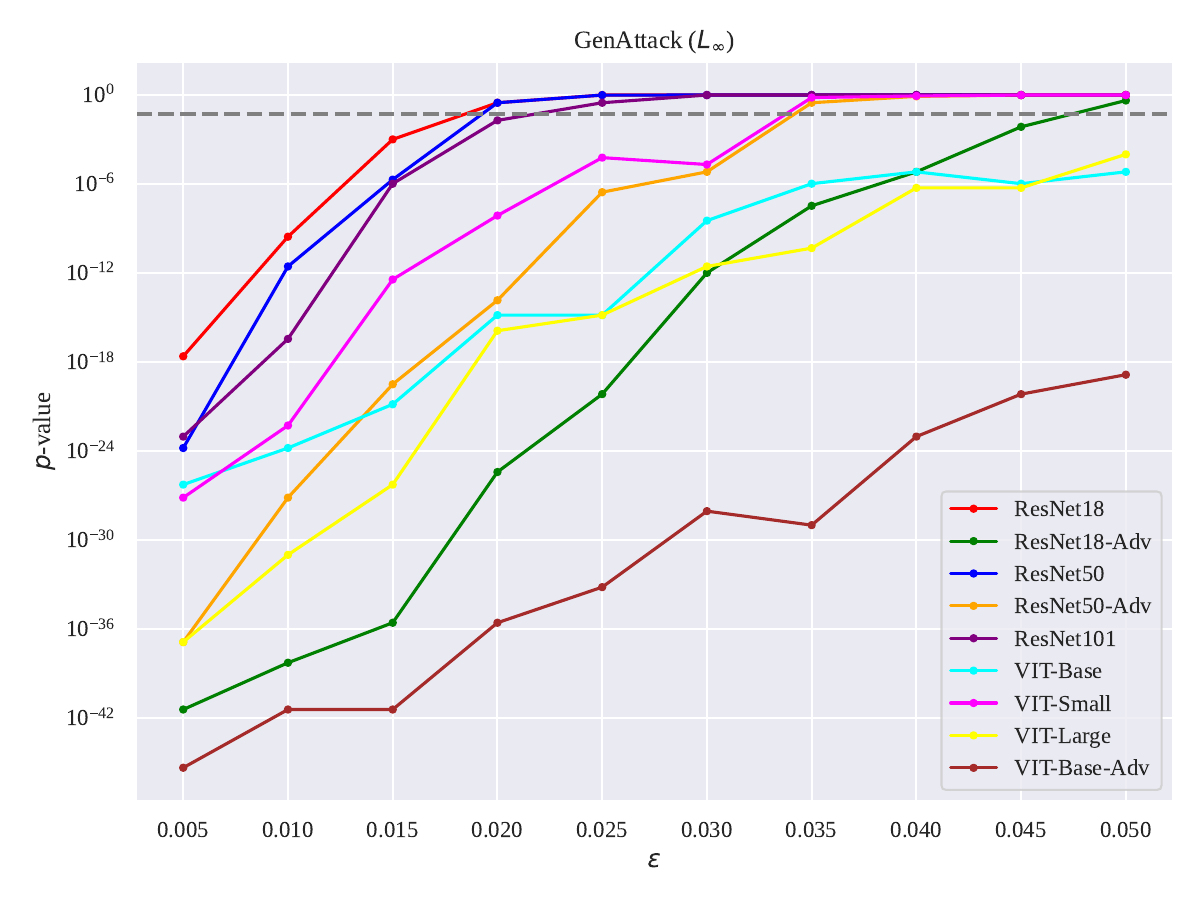}
        \caption{GenAttack with $L_{\infty}$ norm.}
    \end{subfigure}
    \hfill
    \begin{subfigure}[h]{0.33\textwidth}
        \centering
        \includegraphics[width=\textwidth]{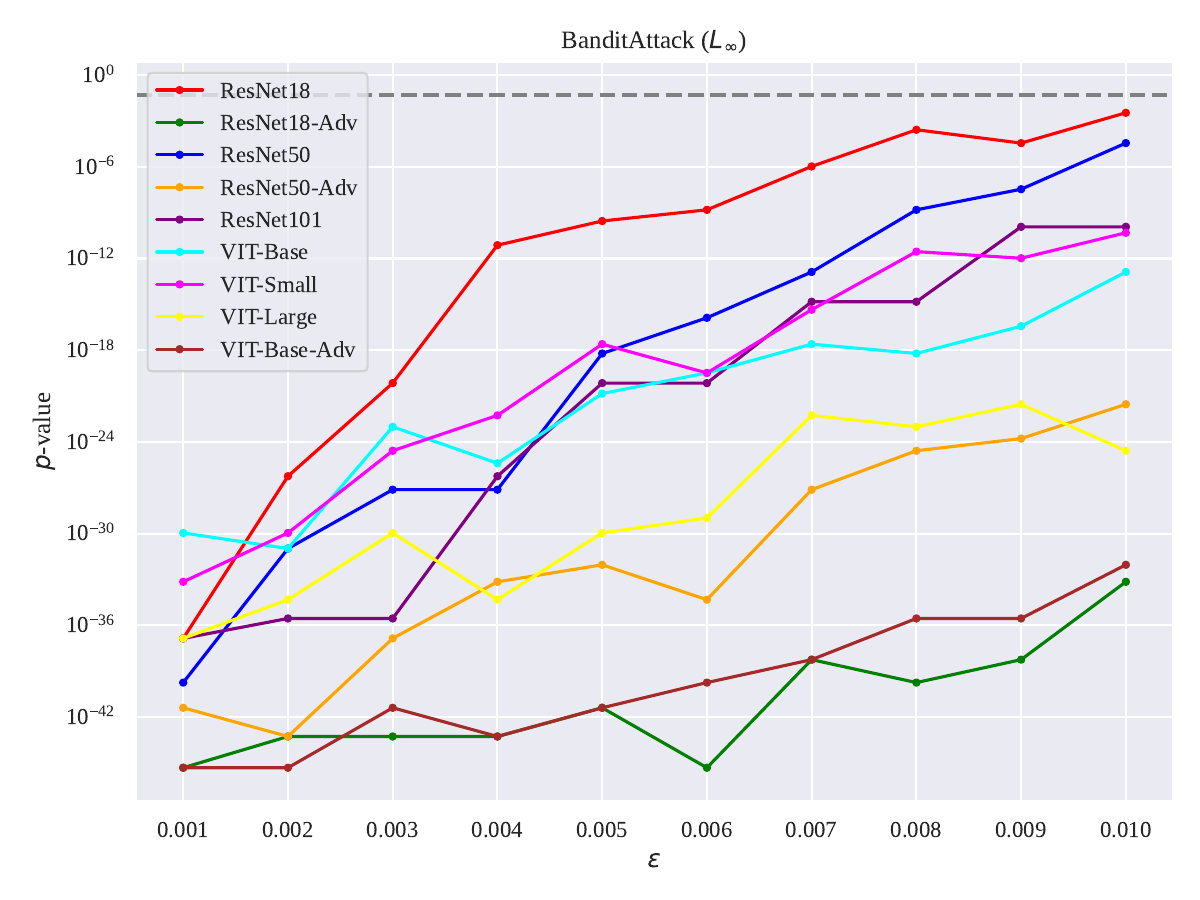}
        \caption{BanditAttack with $L_{\infty}$ norm.}
    \end{subfigure}
    
    \caption{Model certification \textit{w.r.t} different attacking budgets $\epsilon$.}
    \label{fig:all_plots}
\end{figure*}

\begin{theorem}[$(\alpha,\zeta)$-Safe Model with GP-UCB] \label{thm:model safety with gp ucb}
 Fix $0 \leq \alpha \leq 1$, $0 \leq \zeta \leq 1$, $0 \leq \delta \leq 1$ (with $\delta < \zeta$), the machine learning model $\mathcal{M}$, the adversarial attack $\mathcal{A}$ (its budget $\epsilon$ and $\ell_q$-norm). 
Assume that one rejects the null hypothesis provided that GP-UCB $p$-value estimate $\hat{p}_T$ is below $\zeta'$ in \cref{threshold}. Then, one can guarantee that the machine learning model $\mathcal{M}$ is $(\alpha,\zeta)$-safe under an adversarial attack $\mathcal{A}$
for all attack hyper-parameters, i.e.,
    \begin{align}
\mathbb{P} \left({\sf reject}~ \mathcal{R}^* > \alpha ~\big|~ \mathcal{R}^* > \alpha {\sf~is~true} \right) \leq \zeta.
    \end{align}
\end{theorem}

\section{Experiments}

In this section, we conduct extensive experiments with PROSAC to certify the performance of various state-of-the-art vision models in the presence of various adversarial attacks; how the framework recovers existing trends relating to the robustness of different models against different adversarial attacks; and how the framework also suggests new trends relating to state-of-the-art model robustness against attacks.

\subsection{Experimental Settings}
\subsubsection{Datasets} We will consider primarily classification tasks on the ImageNet-1k dataset \citep{deng2009imagenet}. We follow the common experimental setting in black-box adversarial attacks, using 1,000 images from ImageNet-1k \citep{andriushchenko2020square,ilyas2018black} to apply our proposed certification procedure. In particular, we take our calibration set to correspond to this dataset.

\subsubsection{Models} We use two representative state-of-the-art models in computer vision in our experiments, vision transformer (ViT) \citep{dosovitskiy2020image} and ResNet \citep{he2016deep}. We first consider supervised pre-trained models on ImageNet-1k: We use small, base and large models for both ResNet and ViT. Specifically, we test \textit{ViT-Small}, \textit{ViT-Base} and \textit{ViT-Large} for ViT, and \textit{ResNet-18}, \textit{ResNet-50} and \textit{ResNet-101} for ResNet. To certify, we also consider adversarially pre-trained (Adv) models on ImageNet-1k. We adopt the adversarial training models provided by RobustBench, in particular, \textit{ResNet18-Adv}~\citep{salman2020adversarially}, \textit{ResNet50-Adv}~\citep{wongfast} and \textit{ViT-Base-Adv}~\citep{mo2022adversarial}. In summary, we test 9 pre-trained models in our experiments, also see \cref{fig:all_plots}(a) for their classification performance and number of parameters.

\subsubsection{Attackers:} 
We consider both white-box and black-box attackers in our experiments - PGDAttack~\cite{pgdattack} and MomentumAttack~\cite{momentumattack} for white-box, and BanditAttack~\cite{banditattack} and GenAttack~\cite{genattack} for black-box. We default to $L_{\infty}$ perturbation. For comparison, we also consider PGDAttack with $L_2$ perturbation. The hyperparameters of each attacker were carefully selected to explore a wide range of configurations. Specifically, detailed range/values of hyperparameters for each attackers are shown in \textsc{Appendix A}. 
\textit{As our focus is not to investigate and compare different adversarial attackers, for better clarity, we leave it to the interested readers more details and mechanisms about each attacker.} We follow the implementation of different attackers in \texttt{advertorch}~\cite{ding2019advertorch} in our experiments. All hyperparameters can be corresponded back to. 
We set $\alpha=0.10$ and $\zeta=0.05$ in our safety certification procedure, per Definition \ref{def:model safety}.

\subsection{Model safety certification under variable attack budget}

In this section, we certify the safety of nine different models against adversarial attacks, focusing on five different adversarial attack strategies with the hyperparameter range defined earlier, as illustrated in~\cref{fig:all_plots}. As previously discussed, the attack budget $\epsilon$ controls the degree of perturbation applied to the input. We can generally observe that as the attack budget $\epsilon$ increases, the safety of various models decreases (indicated by a rise in the $p$-value). This further explains why we do not consider attack budget $\epsilon$ as a interested hyperparameter - since it has a clear linear relationship with model safety. Additionally, we highlight the following observations:

\begin{itemize}
    \item \textbf{\textit{Large Models vs. Small Models}}: It is commonly agreed that although larger models are capable of capturing complex patterns and generalizing more effectively, they are often more susceptible to overfitting, particularly when trained with limited datasets. This leads to a natural hypothesis that larger models, with more parameters, might be more vulnerable to adversarial attacks, as they could be easily deceived by perturbations that exploit these overfitted details. However, contrary to this hypothesis, we find that smaller models tend to exhibit lower safety against adversarial attacks across both ResNet and ViT model families (i.e., $p^*_{\text{ResNet18}} > p^*_{\text{ResNet50}} > p^*_{\text{ResNet101}}$, $p^*_{\text{ViT-Small}} > p^*_{\text{ViT-Base}} > p^*_{\text{ViT-Large}}$). This counterintuitive result suggests that the relationship between model size and adversarial robustness is more complex than initially assumed, and it highlights the need for further exploration and analysis in this area.
    \item \textbf{\textit{ResNet vs. ViT}}: In addition to the intriguing relationship between model size and model security, we observe a notable contrast between ResNet-family models and ViT-family models - that the ViT-family consistently demonstrates superior adversarial safety compared to the ResNet-family. While a detailed investigation of this phenomenon is beyond the scope of this paper, one key difference lies in their architectural approaches. Unlike ResNet, which relies on local convolutions for feature extraction, Vision Transformers (ViT) utilize a self-attention mechanism. This mechanism allows ViTs to capture long-range dependencies and global context more effectively. By focusing on global interactions between patches of the input image, ViTs may potentially counteract adversarial perturbations that exploit local features.
\begin{figure*}[htbp]
\centering
\begin{subfigure}[b]{0.33\linewidth}
\centering
\includegraphics[width=\textwidth]{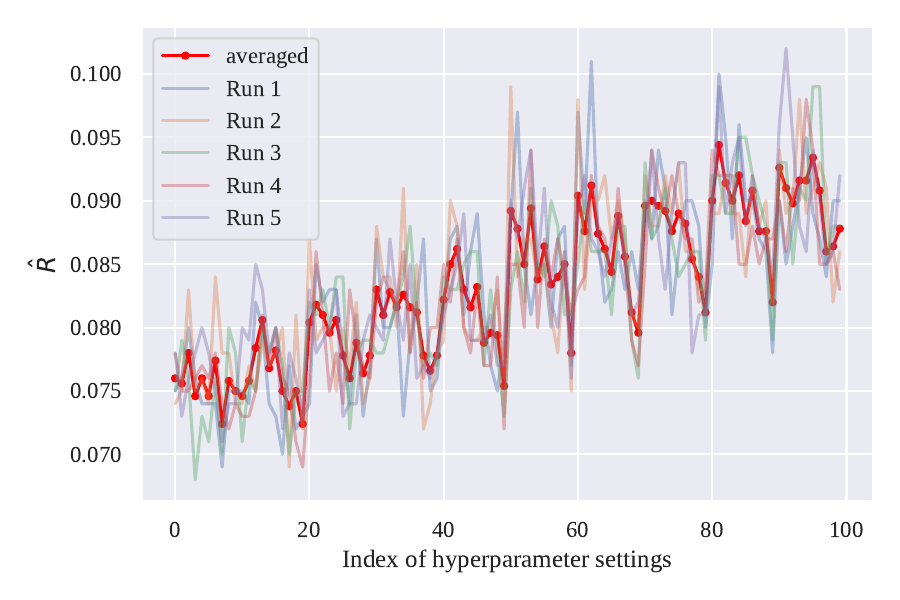}
\caption{Grid search of $\hat{R}$.}
\end{subfigure}
\begin{subfigure}[b]{0.33\linewidth}
\centering
\includegraphics[width=\textwidth]{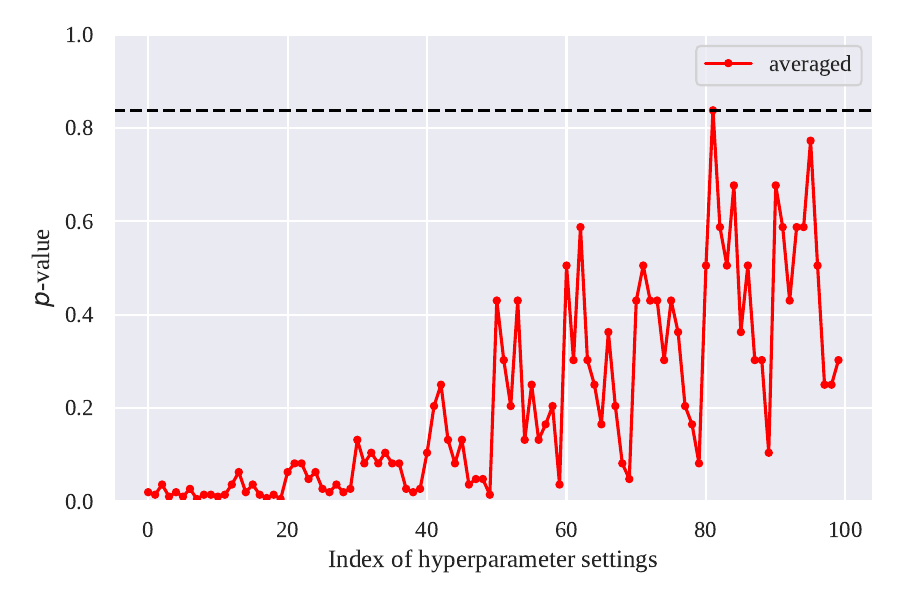}
\caption{Estimated $p$-value with averaged $\hat{R}$.}
\end{subfigure}
\begin{subfigure}[b]{0.33\linewidth}
\centering
\includegraphics[width=\textwidth]{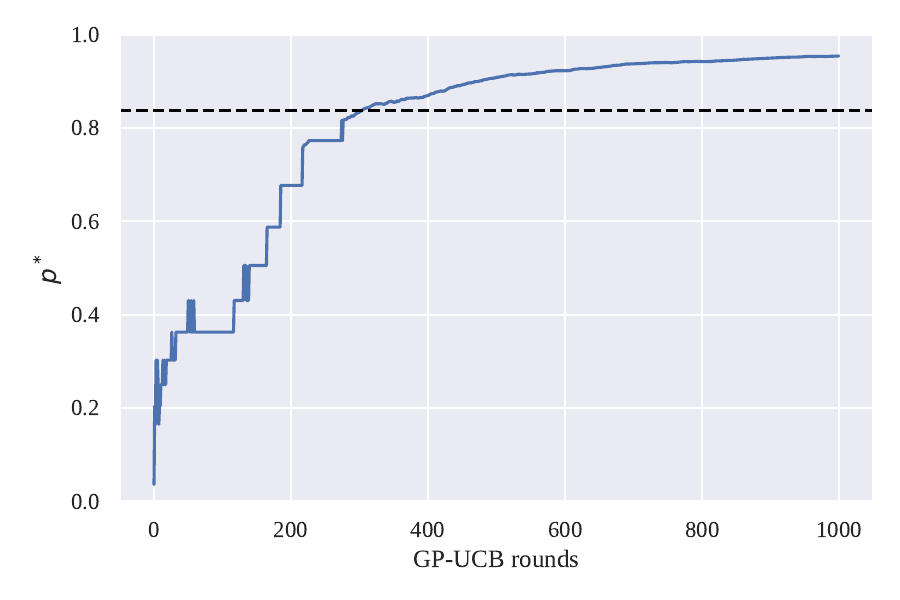}
\caption{GP-UCB estimation.}
\end{subfigure}
\caption{Performance of GP-UCB estimation as the rounds change.}
\label{fig:gpucbestimation}
\end{figure*}

    \item \textbf{\textit{Adversarial Training}}: We also evaluate the performance of existing adversarial training methods. As expected, models incorporating adversarial training techniques (ResNet18-Adv, ResNet50-Adv and ViT-Base-Adv) generally exhibit improved adversarial safety. However, it is important to note that these models also experience a significant impact on their classification performance, see \cref{fig:all_plots}(a). The trade-off between enhanced adversarial resilience and reduced classification accuracy must be carefully considered when implementing adversarial training.
    \item \textbf{\textit{Different Attackers}}: Finally, we conducted a comparison of different attackers. Note that, since the attack budget $\epsilon$ heavily influences the attacker's capability, we only provide a rough summary based on significant observations available. It is noted that, compared to black-box attacks, white-box attacks are more harmful even with lower attack budgets. We conjecture that, in white-box attacks, the attacker has complete access to the model's internal parameters and gradient information, enabling the design of more effective adversarial samples. 
\end{itemize}

\subsection{Approximation performance of GP-UCB}
Before the primary model validation process, we first empirically validated the performance of the GP-UCB method in estimating the $p$-value. Specifically, this section focuses on evaluating the robustness of the ResNet18 model when subjected to the GenAttack. For comparison, in addition to the proposed GP-UCB method for estimating the $p$-value, a straightforward grid search approach was also employed. To ensure computational efficiency, we only considered two variable parameters
, resulting in a total of 100 hyperparameter combinations. 

As shown in \cref{fig:gpucbestimation}(a), we present the \textit{attack success rate} $\hat{R}$ corresponding to these 100 parameter combinations. The final attack success rate (`averaged') was obtained by averaging the results from five random repetitions (`Run 1' - `Run 5'). Based on \cref{eqn:p-value} and \cref{eq:attacksuccessrate}, we calculated the $p$-values corresponding to each hyperparameter combination, shown in \cref{fig:gpucbestimation}(b). Notably, we marked the maximum value ($p^*$) with a horizontal line. For comparison, \cref{fig:gpucbestimation}(c) illustrates the $p^*$ values estimated by the GP-UCB algorithm at different rounds. We observed that, compared to grid search, after a certain number of iterations, the GP-UCB method consistently provides a more conservative $p$-value ($p^*_{gpu-cb} > p^*_{grid-search}$), which helps further reduce the false positive rate in the model robustness assessment.

\section{Conclusions}
We have proposed PROSAC, a new approach to certify the performance of a machine learning model in the presence of an adversarial attack, with population level adversarial risk guarantees. PROSAC builds on recent work on distribution-free risk quantification approaches, offering an instrument to ascertain whether a model is likely to be safe in the presence of an adversarial attack, independently of how the attacker chooses the attack hyperparameters. We show via experiments that PROSAC is able to certify various state-of-the-art models, leading to results that are in line with existing results in the literature. 

The technical framework developed here is likely to be of high relevance to AI regulation, such as the EU's AI Act, which requires providers of certain AI systems to ensure that their systems are resilient to adversarial attacks. Our approach to certifying the performance of any black-box machine learning system offers a tool that can help providers to comply with their legal obligations.
Beyond its utility as a certification instrument, our framework also suggests that a number of areas in adversarial robustness may merit further attention. First, PROSAC has shown that large ViT models appear to be more adversarially robust than smaller models, pointing to new directions for research on the relationship between the capacity of a ViT and its adversarial robustness. Second, CLIP-ViT models appear to be more vulnerable to adversarial attacks than supervised trained ViTs, raising questions about how to use self-supervised models better in improving adversarial robustness of downstream tasks. 

We highlight the importance of measuring the resilience of ML models against adversarial use and misuse, not only in order to comply with the EU AI Act, but also because general duties to act with due care and not negligently presuppose that providers of AI systems have clarity about the safety of their models. Both the AI Act and the duty of care are intended to protect society against harms caused by malfunctioning AI systems. With our approach to certifying robustness, we seek to provide a tool to reduce these societal risks. One limitation of our approach is that the framework can only control the Type-I error and is not able to control the Type-II error, i.e., the probability of declaring that the model is not safe when the model is in fact safe. Therefore, our framework can be over-conservative in the adversarial safety certification. 
In addition, in future work, we aim to further investigate the security challenges of models in multimodal learning, particularly in the presence of noise and missing modalities~\citep{Feng_2022_BMVC,Feng_2024_ACMMM,zhi2024borrowing,zhi2024wasserstein}.
However, we believe that as a first step towards safety certification, it is not necessarily negative to be too cautious. 

\section{Acknowledgements}
We acknowledge support from Leverhulme Trust via research grant RPG-2022-198.

\bibliography{aaai25}

\end{document}